\documentclass[10pt,twocolumn,letterpaper]{article}

\usepackage{cvpr}
\usepackage{times}
\usepackage{graphicx}
\usepackage{amsmath}
\usepackage{amssymb}
\usepackage{caption}
\usepackage{cuted}
\usepackage[numbers,sort&compress]{natbib}
\usepackage{booktabs}
\usepackage{url}

\cvprfinalcopy

\begin{document}

\title{
}

\title{When Reality Outpaces Cognition:\\
\Large\normalfont\textit{A New Scaling Principle for Reliable Intelligence}}

\author{
\textbf{Ahsan Adeel}$^{1}$\thanks{$^{1}$University of Stirling \\ Correspondence: 
\texttt{ahsan.adeel1@stir.ac.uk}}
\qquad
\textbf{Talha Bin Riaz}$^{1}$\\[1pt]
TREND Project $|$ Funded by the Advanced Research + Invention Agency (ARIA)
}

\maketitle
\vspace{-6pt} % Reduce gap between authors and figure
\begin{strip}
\vspace{-35pt}
    \centering
    \includegraphics[width=1\textwidth]{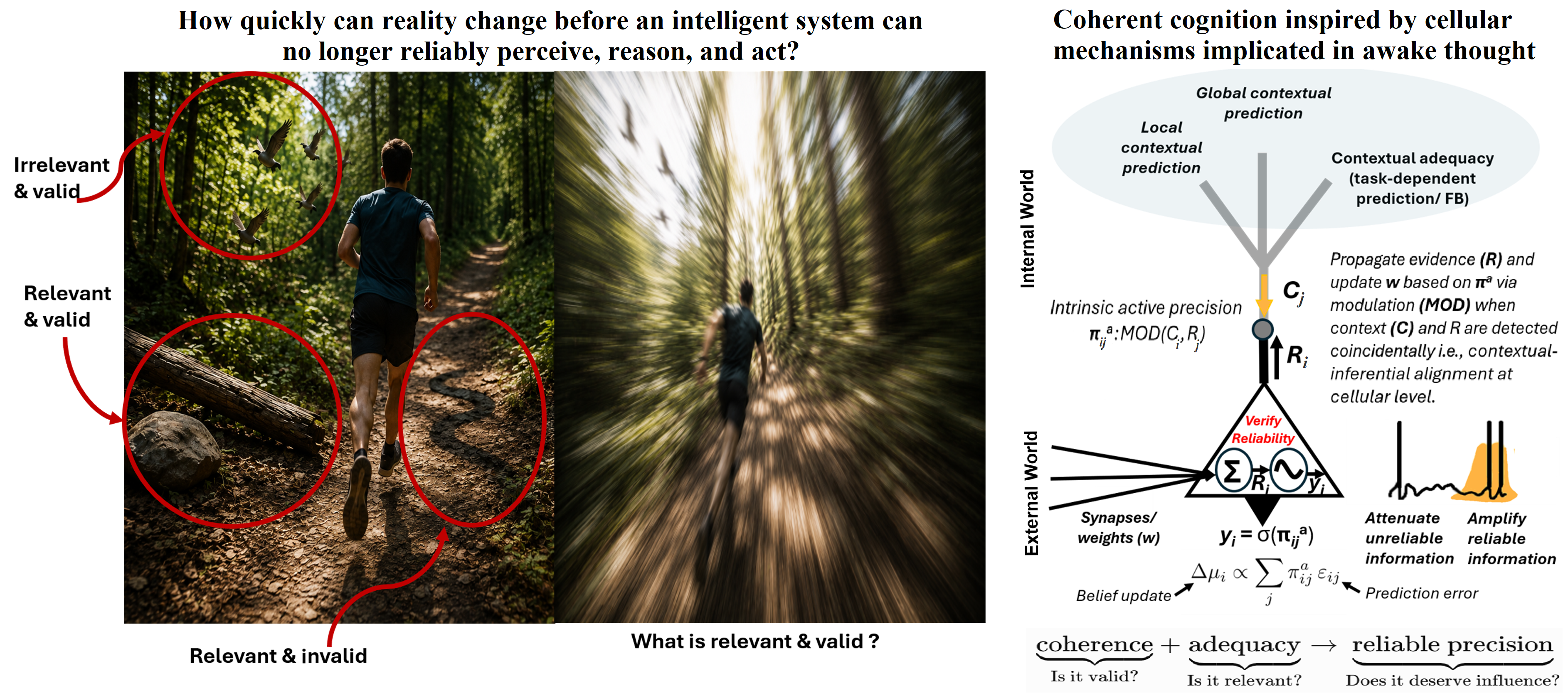}
\captionof{figure}{\textbf{Left:} To probe whether AI models  understand the world rather than merely predict what comes next, challenge them to infer the consequences of actions for far-future states they have not yet observed, under ambiguous and rapidly changing conditions. Consider reality evolving through world states, State 1 $\rightarrow$ State 2 $\rightarrow \cdots \rightarrow$ State 50.
An agent has experienced these states successively at the
Speed of Reality (\(S_R=1\)), but at \(50\times\) Speed of Reality
(\(S_R=50\)), reality evolves by a factor of 50 across the agent's
observation--action cycle, creating a State 1 $\rightarrow$ State 50
temporal relationship that was not experienced during training. Can it infer
from State 1 an action appropriate for State 50 before State 50 is
observed? As the observation--reality gap grows, increasing visual blur
represents growing ambiguity and difficulty in maintaining reliable
perception, reasoning, and action, not literal sensory blur.
\textbf{Right:} We hypothesize that architectures such as Cooperative Context-sensitive Cognitive Computation (\(Co^4\)) \cite{adeel2026scalable,adeel2025beyond}, inspired by cellular mechanisms implicated in conscious processing \cite{aru2020cellular,storm2024integrative} and awake thought \cite{Phillips2024cellular, graham2026inhibitory}, provide one possible route to addressing the problem of relevance \cite{parvizi2025active} and scaling coherent cognition beyond a fundamental limitation shared by biological and artificial systems when outpaced by reality.}
    \label{fig:forest}
\end{strip}

\begin{figure*}
    \centering
    \includegraphics[
        trim=0cm 0cm 0cm 0cm,
        clip=true,
        width=0.95\linewidth
    ]{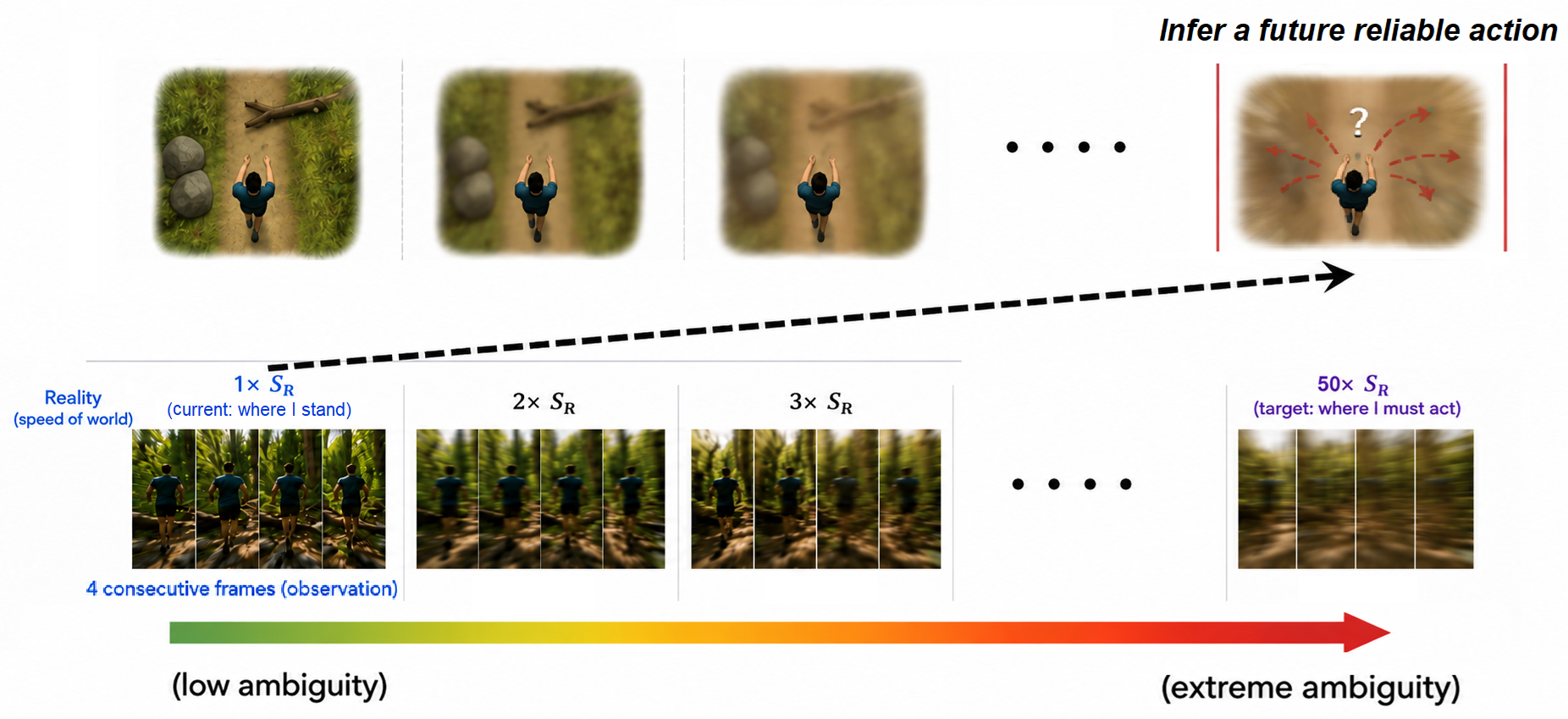}
\caption{
\textbf{Operational challenge of when reality outpaces cognition.}
The agent is trained at the baseline Speed of Reality (\(S_{R,0}=1\)),
but tested at \(50\times\) the baseline (\(S_R=50\)) while computational
resources remain fixed. From State 1, it must infer the action appropriate
for the future State 50 while States 2--50 remain unobserved. The challenge
is whether the agent has learned sufficiently generalisable task-relevant
world dynamics to infer a reliable action across this observation--reality gap.
}
    \label{fig:test}
\end{figure*}
\textit{
\textbf{Abstract}: This paper challenges the understanding capabilities of existing AI paradigms and emerging world models by asking a fundamental question: How quickly can reality change before an intelligent system can no longer reliably perceive, reason, and act?~We formulate this as a theory- and model-agnostic challenge in which systems must infer reliable actions for increasingly distant future states across a growing observation--reality gap. We operationalize this challenge through the effective Speed of Coherent Cognition (\(S_C^{\mathrm{eff}}\)), defined as the highest tested Speed of Reality (\(S_R\)) through which a system continuously satisfies a predefined behavioural reliability criterion from baseline. As a proof of concept, reinforcement-learning agents are trained at the baseline \(S_R\) and evaluated, without retraining, as \(S_R\) is increased by up to \(50\times\) while computational resources remain fixed. We compare Transformer and gated-MLP baselines with \(Co^4\), an architecture inspired by cellular mechanisms implicated in conscious processing and awake thought.~Across the tested scenarios, \(Co^4\) achieves substantially higher \(S_C^{\mathrm{eff}}\) than the baselines, motivating \(S_C^{\mathrm{eff}}\) as a new benchmark and the scaling of coherent cognition as a principle for reliable AI.
}

\section{Introduction}
Increasing $S_R$ does not merely evaluate the processing
speed of an intelligent system, but rather its ability to
infer the consequences of actions and maintain reliable
behaviour in rapidly changing situations that have not
yet been directly observed. One might ask why an AI system should not simply be trained at the \(S_R\) expected during deployment. However, for future autonomous AI systems, particularly in mission- or safety-critical settings, it may not be possible to guarantee that training exhaustively captures the maximum rate of relevant change, ambiguity, uncertainty, or abrupt environmental dynamics that may be encountered during deployment. Such systems must perceive, reason, and act reliably as environmental dynamics depart from those experienced during training. Understanding where and how this reliability degrades is therefore important for establishing safe operating limits and developing AI systems that remain dependable under unforeseen deployment conditions \cite{dalrymple2024towards}. The proposed challenge is domain-general in formulation and may extend to settings in which physical, informational, multi-agent, or adversarial dynamics outpace a system's capacity to maintain coherent understanding and reliable behaviour, with implications for the development and evaluation of safe and reliable AI. The relationship between the proposed challenge and neighbouring work on
temporal out-of-distribution (OOD) generalisation, frame skipping/action repetition, observation
latency, delayed control, partial observability, dynamics shifts, and world models is discussed in Appendix A
\cite{sharma2017learning, biedenkapp2021temporl, tallec2019making,
ramstedt2019real, ramstedt2020reinforcement, anokhin2025handling,
tessler2019action, zhang2026world, li2019robust, padakandla2021survey,
kirk2023survey, xiao2020thinking, zhang2020robust, tobin2017domain}.

\section{Formalizing $S_R$ and $S_C^{\mathrm{eff}}$}
We begin with an illustrative example of reality outpacing cognition.
Imagine running through a dense forest (Figures 1 and 2). From the
observable environment detected by biological and extended sensors,
internally generated predictions guide attention moment-by-moment toward
contextually valid and relevant signals, such as trees, branches, and rocks,
in order to support reliable action. As bodily motion increases, the local
observable environment changes more rapidly relative to the observer. The
speed of this change determines how fast the cognitive system must update and
use its internal representation of the environment to maintain coherent
understanding and make reliable and useful probabilistic judgments. At
ordinary speeds, the runner observes the environment evolving through
successive situations. For example, an approaching obstacle may be observed
as \textit{$\text{far} \rightarrow \text{closer} \rightarrow \text{close}
\rightarrow \text{very close}$} allowing the runner to continuously update action as the situation evolves.

Now imagine that the speed of this observable reality increases beyond the
rate at which the cognitive system can maintain and use such coherent
representations. As the runner continues, the forest rushes past faster and
faster. In the extreme case, for example at \(S_R=50\), the runner observes
the forest at State 1 and must infer the action appropriate for the future
State 50 while States 2--50 remain unobserved. For example, an approaching
obstacle may be observed as \textit{$\text{far} \rightarrow \text{very close}$},
without the intermediate states being observed. Rather than observing and updating action across the successive states
experienced at ordinary speeds, the runner must bridge an increasing
observation--reality gap. As more unobserved environmental changes occur, uncertainty about the future ground-truth state increases. Trees, branches, and rocks become increasingly difficult to distinguish in
time, while previously adequate interpretations may become outdated before
they can reliably guide action. Increasing visual blur (Figure 1, left) is a
metaphor for the growing ambiguity caused by unobserved environmental change,
not literal sensory blur.

Eventually, environmental dynamics exceed the system's capacity to integrate
available evidence, maintain a coherent and useful representation of its
surroundings, and infer an appropriate action for the relevant future state
before that state is observed. The resulting mismatch increases the likelihood
of misjudgment and harmful action, such as colliding with a tree or stumbling
over a rock. We refer to this regime as \textit{mind--reality\footnote{A conceptual label for the relationship
between the rate of environmental change and a system's capacity for coherent
cognition, without implying any particular account of machine mentality.} overload}.
Conceptually,

\begin{equation}
S_R > S_C
\label{eq:overload}
\end{equation}

where,

\begin{itemize}
\item $S_R$ denotes the \textit{Speed of Reality}, characterizing the rate of task-relevant environmental change relative to the system's observation--action cycle. Increasing $S_R$ therefore increases the amount of environmental evolution occurring across this cycle, enlarging the gap between the currently observed state and the future state for which an action must remain appropriate.

\item $S_C$ denotes the \textit{Speed of Coherent Cognition}, defined as
the rate at which the system can distinguish valid and relevant information
from invalid or irrelevant information, and selectively utilize the former
to maintain reliable perception, reasoning, and action.
\end{itemize}

Importantly, increasing $S_R$ during testing systematically increases the task-relevant environmental change occurring between observation and required action, thereby progressively enlarging the observation–reality gap. However, the broader challenge
is not restricted to temporal acceleration and can be
generalised to other forms of increasing environmental
ambiguity, uncertainty, and complexity that challenge a
system's ability to distinguish valid and relevant information
from invalid or irrelevant information and maintain reliable
behaviour. Thus, $S_R$ provides one operational instantiation
of a more general challenge: determining the limits of
coherent cognition under increasingly demanding
environmental conditions.

The relationship between $S_R$ and $S_C$ can then be expressed through the dimensionless mind--reality ratio:
\begin{equation}
\gamma = \frac{S_C}{S_R}
\label{eq:gamma}
\end{equation}
such that
\begin{align}
\gamma \geq 1
&\quad \text{Mind--reality synchrony} \\
\gamma < 1
&\quad \text{Mind--reality overload}
\end{align}
\\Here, $S_C$ denotes a theoretical capacity, distinct from the empirically
estimated quantity $S_C^{\mathrm{eff}}$ introduced below.

The challenge is not whether a system can process familiar observations faster, but whether its learned task-relevant world dynamics can support reliable action over temporal horizons that increasingly extend beyond those encountered during training.

Let $\mathcal{R}(S_R)$ denote a task-specific behavioural reliability measure
at a given $S_R$, and let $\tau$ denote a predefined minimum reliability
threshold. Reliable behaviour is maintained when:

\begin{equation}
\mathcal{R}(S_R) \geq \tau
\label{eq:reliability}
\end{equation}

Let the finite set of $S_R$ values evaluated experimentally be

\begin{equation}
\mathcal{S}_{\mathrm{test}}
=
\{s_1,s_2,\ldots,s_n\},
\qquad
s_1=S_{R,0}<s_2<\cdots<s_n
\end{equation}

where $s_k$ denotes the $k$th tested value of $S_R$, ordered from the
training baseline. We define the experimentally estimated $S_C^{\mathrm{eff}}$ as

\begin{equation}
S_C^{\mathrm{eff}}
=
\max
\left\{
s_k \in \mathcal{S}_{\mathrm{test}}
\;\middle|\;
\mathcal{R}(s_i)\geq\tau
\;\; \forall\, i\leq k
\right\}
\label{eq:effective_sc}
\end{equation}

$S_C^{\mathrm{eff}}$ is the highest tested $S_R$ through which the
predefined reliability criterion is continuously satisfied from the training
baseline. It therefore represents an empirical behavioural boundary rather
than a direct measure of cognitive speed. Comparisons of
$S_C^{\mathrm{eff}}$ should be made under matched experimental conditions.

\(S_C^{\mathrm{eff}}\) is not an exhaustive measure of understanding, but
characterizes the dynamic reliability boundary at which this capacity begins
to break down. Conceptually, the condition for coherent cognition can be
expressed as
\begin{equation}
S_C \geq S_R
\label{eq:scaling}
\end{equation}

Scaling \(S_C\) therefore means extending the range of environmental dynamics
over which this condition can be maintained. It is not merely a matter of
processing observations faster, but of effectively using available
information and computational resources to maintain task-relevant
representations and select reliable actions as environmental dynamics
accelerate. This motivates the new scaling principle:

\begin{quote}
\textit{Scale $S_C$, not merely FLOPs, parameters, and data.}
\end{quote}

% \begin{figure}[t]
% \centering
% \includegraphics[
%     width=0.9999\linewidth
% ]{co4n}
% \caption{
% }
% \label{fig:co4_cellular}
% \end{figure}

\section{Operationalising $S_C^{\mathrm{eff}}$}
For reinforcement-learning (RL) tasks, Co$^4$ (Figure 1, right)
\cite{adeel2025beyond,adeel2026scalable} is compared with the
Transformer-based architecture proposed in \cite{tang2021sensory} and a
gated MLP (gMLP) \cite{liu2021pay}. We additionally compare Co$^4$ variants in which the form  of
apical modulation \cite{adeel2026scalable, graham2025context, Phillips2024cellular}, including apical amplification (AA) and apical driven + awake (ADA) are systematically altered. The architectures are evaluated under matched experimental conditions, including the same environments and computational constraints, with architecture-specific configurations reported below. As a proof of concept, we evaluate the proposed test in two environments:
car racing \cite{tang2021sensory, ha2018world} and complex vision-guided quadrupedal locomotion \cite{yang2021learning}. 

\subsection{The \(Co^4\) Computational Principle}
$Co^4$ evaluates incoming evidence against local and global
contextual predictions and contextual adequacy via modulatory
(MOD) functions \cite{adeel2026scalable, graham2025context}
before substantially influencing downstream processing.
Evidence that is both valid and relevant is amplified
(effectively assigned greater precision
\cite{friston2010free}), whereas poorly supported evidence
is attenuated, implementing the principle of
\textit{verification before propagation}.

In the proof-of-concept experiments considered here,
$Co^4$ implements this principle through triadic modulation
loops among $Q$-, $K$-, and $V$-two-point neurons (TPNs)
\cite{adeel2026scalable}, which modulate information
according to contextual support before attentional
competition. For $Q$-TPNs, each token provides a primary
evidence stream ($R$), while local and global contextual
inputs ($C$) are provided by the $K$ and $V$ projections,
respectively. The same principle applies to $K$- and
$V$-TPNs. Adequacy represents task-dependent prediction
or feedback. Further details are provided in
\cite{adeel2026scalable,adeel2026cellular}.

\subsection{Vision-Guided Quadrupedal Locomotion}
We evaluate vision-guided quadrupedal locomotion under two task conditions
of increasing complexity: easy and complex. Here, “easy” and “complex” refer to scenarios in which objects are farther from and closer to the agent, respectively. All systems are trained only at the baseline Speed
of Reality, $S_{R,0}=1$, and subsequently evaluated without retraining as
$S_R$ is increased while computational resources remain fixed. Because the
two conditions differ in task difficulty and reward scale, reliability
thresholds are predefined separately for each condition.
\subsubsection{Easy Locomotion Task}
In the easier vision-guided locomotion environment, where obstacles are more widely separated, all architectures learn effective policies under the baseline training condition (\(S_R=1\)), although with different learning curves and asymptotic rewards (Figure~3). The Transformer and $Co^4$ architectures each consist of two layers and four attention heads, whereas the gMLP architecture consists of two layers. The Transformer achieves a baseline performance comparable to the \(Co^4\) variants, whereas the gMLP converges to a lower reward. This environment therefore provides a useful setting for testing whether policies that perform successfully under the training condition differ in their ability to maintain reliable behaviour as \(S_R\) increases.
\begin{figure}
\centering
\includegraphics[
    trim=0cm 0cm 0cm 0cm,
    clip=true,
    width=1\linewidth
]{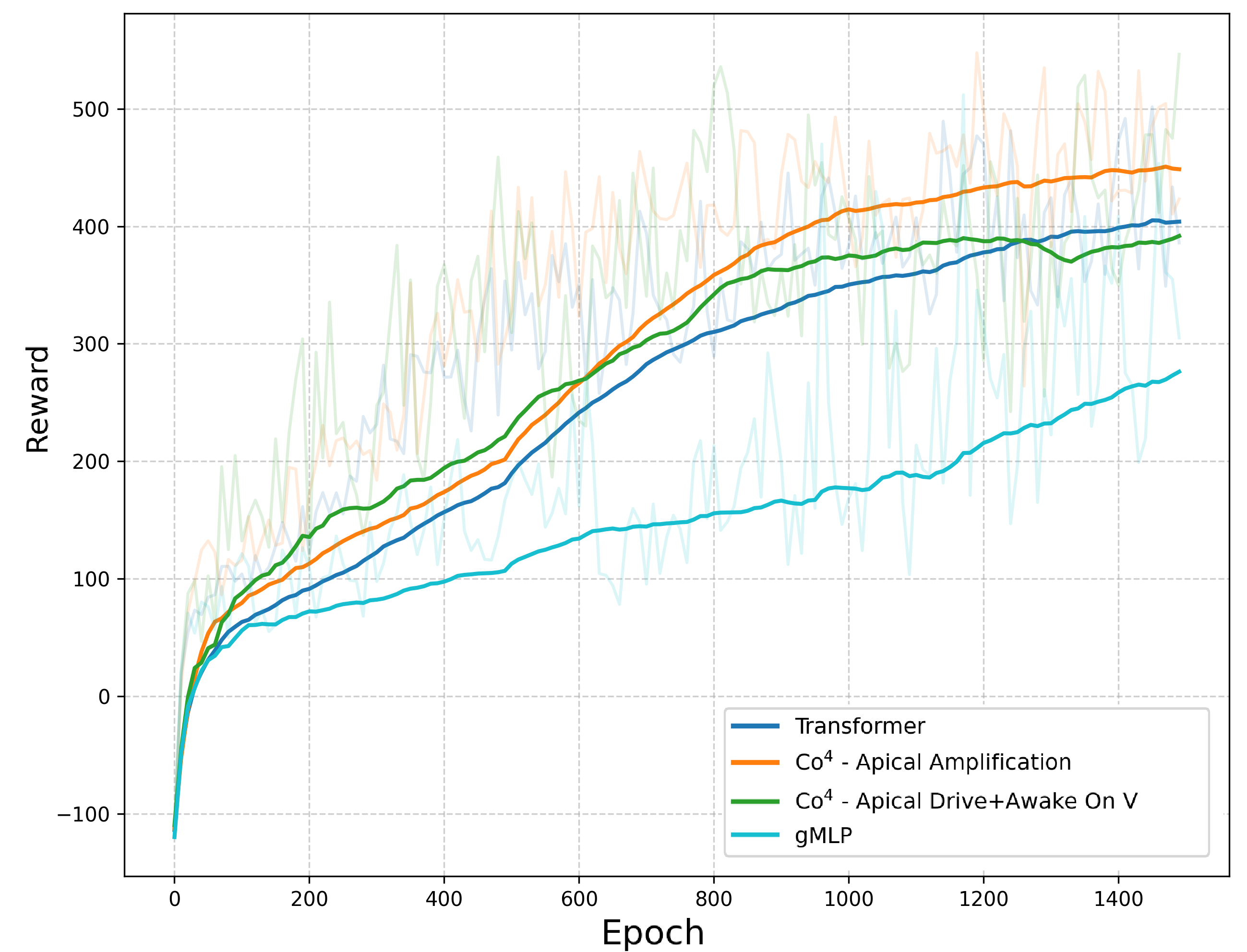}
\caption{\textbf{Training performance in the easy vision-guided quadrupedal locomotion environment.}
Running average reward over training epochs for the Transformer, gMLP, and two $Co^4$ variants at $S_{R,0}=1$. In the easy environment, obstacles are more widely separated, allowing all architectures to learn the locomotion task, although with different learning trajectories and final performance levels.}
\label{fig:co4_ablation}
\end{figure}
\begin{figure} 
\centering
\includegraphics[
    trim=0cm 0cm 0cm 0cm,
    clip=true,
    width=1\linewidth
]{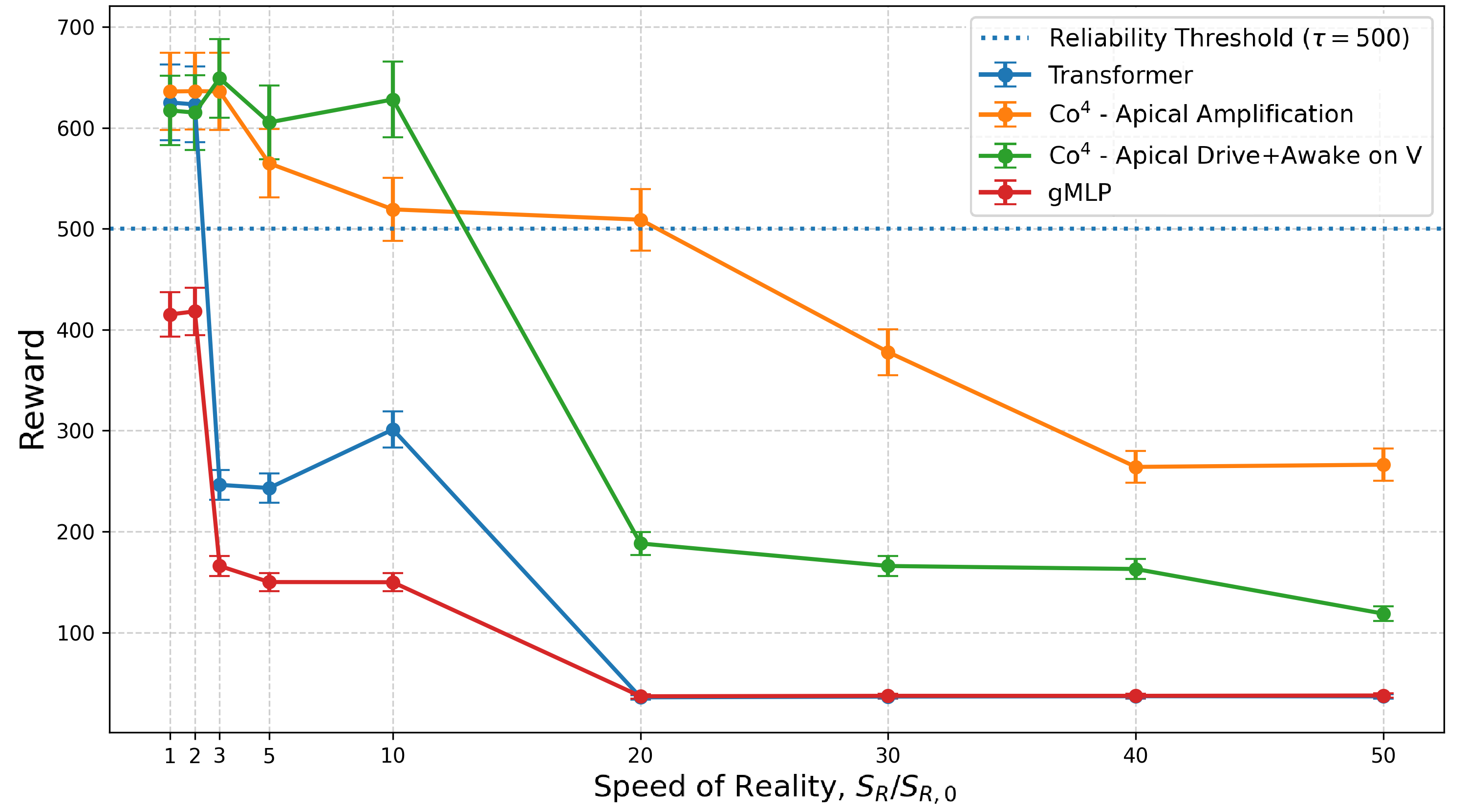}
\caption{
\textbf{{Scaling $S_C^{\mathrm{eff}}$ in the easy vision-guided quadrupedal locomotion task.}}
Reward is shown as $S_R$ increases from the training baseline
$S_{R,0}=1$ to $S_R=50$ without retraining. The dashed line denotes the
predefined reliability threshold $\tau=500$.}
\label{fig:reward_graph_randomshape_4_whiskers 1}
\end{figure}
\begin{table}[t]
\centering
\caption{Performance in the easy vision-guided quadrupedal locomotion task as $S_R$ increases. Values are mean $\pm$ standard deviation.}

\label{tab:easy_locomotion}
\resizebox{\columnwidth}{!}{%
\begin{tabular}{ccccc}
\toprule
$\mathbf{S_R}$ 
& $\mathbf{Co^4}$-AA 
& $\mathbf{Co^4}$-ADA 
& \textbf{Transformer} 
& \textbf{gMLP} \\
\midrule
1  & 636.0 $\pm$ 38.2 & 617.3 $\pm$ 34.2 & 625.2 $\pm$ 37.5 & 415.1 $\pm$ 22.1 \\
2  & 636.4 $\pm$ 38.2 & 615.1 $\pm$ 36.9 & 623.2 $\pm$ 37.4 & 418.2 $\pm$ 23.6 \\
3  & 636.1 $\pm$ 38.2 & 649.2 $\pm$ 39.0 & 246.3 $\pm$ 14.8 & 166.0 $\pm$ 10.0 \\
5  & 564.9 $\pm$ 33.9 & 605.4 $\pm$ 36.3 & 243.2 $\pm$ 14.6 & 150.0 $\pm$ 9.0 \\
10 & 519.3 $\pm$ 31.2 & 628.2 $\pm$ 37.7 & 301.2 $\pm$ 18.1 & 149.8 $\pm$ 9.0 \\
20 & 509.0 $\pm$ 30.5 & 188.3 $\pm$ 11.3 & 35.8 $\pm$ 2.1   & 36.9 $\pm$ 2.2 \\
30 & 377.8 $\pm$ 22.7 & 165.9 $\pm$ 10.0 & 36.6 $\pm$ 2.2   & 37.3 $\pm$ 2.3 \\
40 & 264.1 $\pm$ 15.8 & 163.0 $\pm$ 9.8  & 36.9 $\pm$ 2.2   & 37.3 $\pm$ 2.2 \\
50 & 266.3 $\pm$ 16.0 & 118.7 $\pm$ 7.1  & 36.8 $\pm$ 2.2   & 37.6 $\pm$ 2.3 \\
\bottomrule
\end{tabular}%
}
\end{table}
Figure~4 and
Table~1 show the resulting reward as $S_R$
increases without retraining. The Transformer satisfies the reliability criterion through
$S_R=2$ and first falls below the threshold at $S_R=3$, yielding
$S_{C,\mathrm{Transformer}}^{\mathrm{eff}}=2$. Co$^4$-ADA maintains
mean reward above the reliability threshold through $S_R=10$ and falls
below the threshold at $S_R=20$, yielding
$S_{C,\mathrm{Co}^4\text{-ADA}}^{\mathrm{eff}}=10$. Co$^4$-AA remains
above the reliability threshold through $S_R=20$ and falls below the
threshold at $S_R=30$, yielding
$S_{C,\mathrm{Co}^4\text{-AA}}^{\mathrm{eff}}=20$. The corresponding
transitions are therefore localized between $S_R=10$ and $S_R=20$ for
Co$^4$-ADA, and between $S_R=20$ and $S_R=30$ for Co$^4$-AA. The
standard gMLP does not satisfy the predefined reliability criterion at
the training baseline, $S_{R,0}=1$, and therefore does not satisfy the
baseline eligibility requirement for estimating $S_C^{\mathrm{eff}}$
under this test; its $S_C^{\mathrm{eff}}$ is left undefined.

Representative attention visualizations provide a qualitative view of how
the architectures process task-relevant visual information.
Figure 5 shows comparatively diffuse attention
for the Transformer baseline and more selective attention for Co$^4$ with
contextual modulation. When the obstacle is distant, Co$^4$ concentrates
attention on the immediately relevant traversable region and shifts attention
toward the obstacle as it becomes locally relevant. The corresponding
sequence-level attention map is also substantially sparser. These examples
are consistent with the hypothesis that contextual modulation restricts the
propagation of currently irrelevant and invalid information before attentional
competition.

\begin{figure}
    \centering
    \includegraphics[
        trim=0cm 0cm 0cm 0cm,
        clip=true,
        width=1\linewidth
    ]{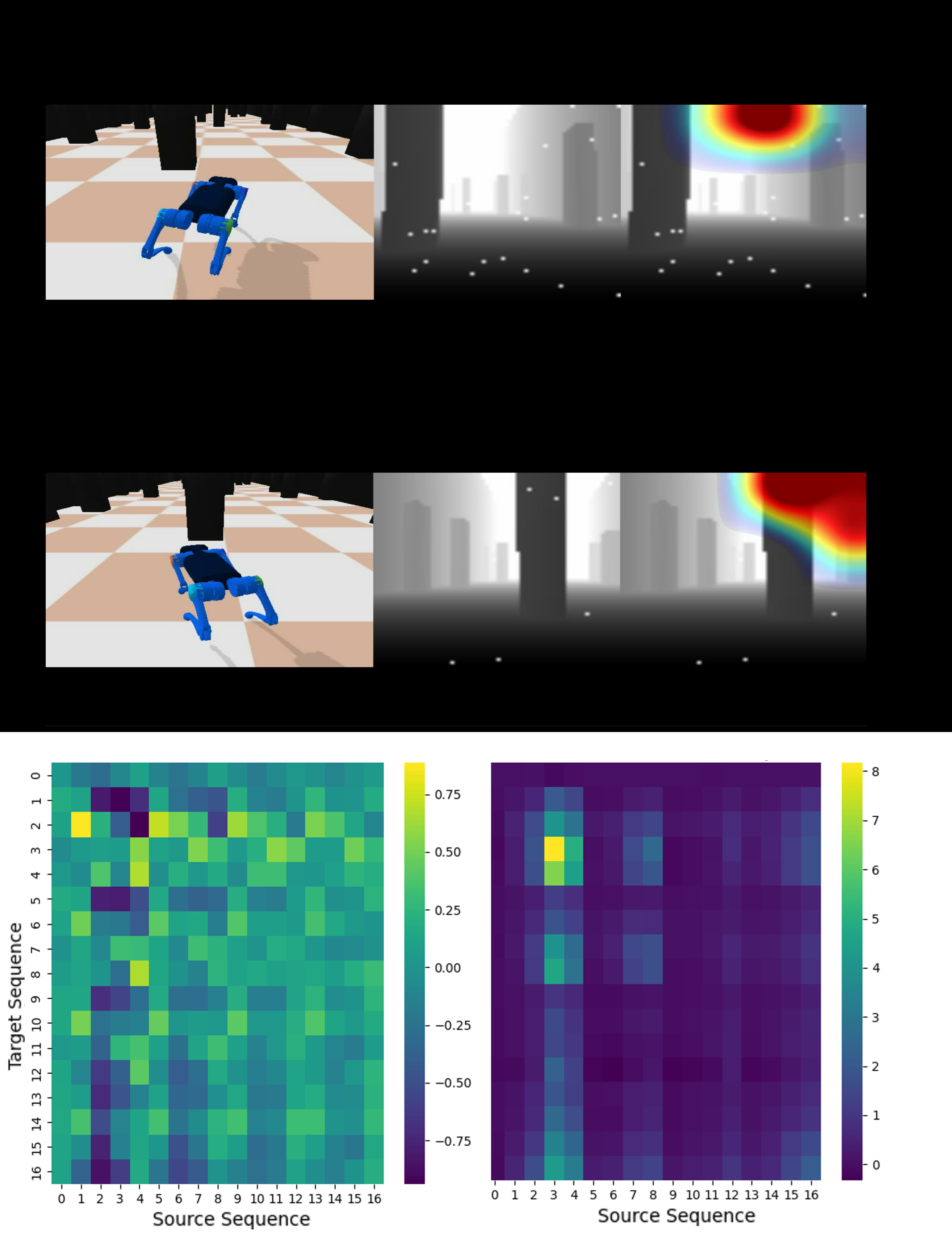}
    \caption{
    \textbf{Representative attention patterns in vision-guided quadrupedal
    locomotion.}
    The Transformer exhibits comparatively diffuse attention (top): when the
    obstacle is distant, attention extends substantially over background
    regions, while proximity to the obstacle produces broadly distributed
    attention across the visual input. In contrast, Co$^4$ (middle) exhibits more
    selective attention, concentrating on the immediately relevant
    traversable region while the obstacle is distant and shifting toward the
    obstacle as it becomes locally relevant. This difference is also
    reflected in the sequence-level attention maps (bottom), where Co$^4$ produces a
    substantially sparser pattern. }
    \label{fig:quadruped_attention}
\end{figure}
\begin{figure}
\centering
\includegraphics[
    trim=0cm 0cm 0cm 0cm,
    clip=true,
    width=1\linewidth
]{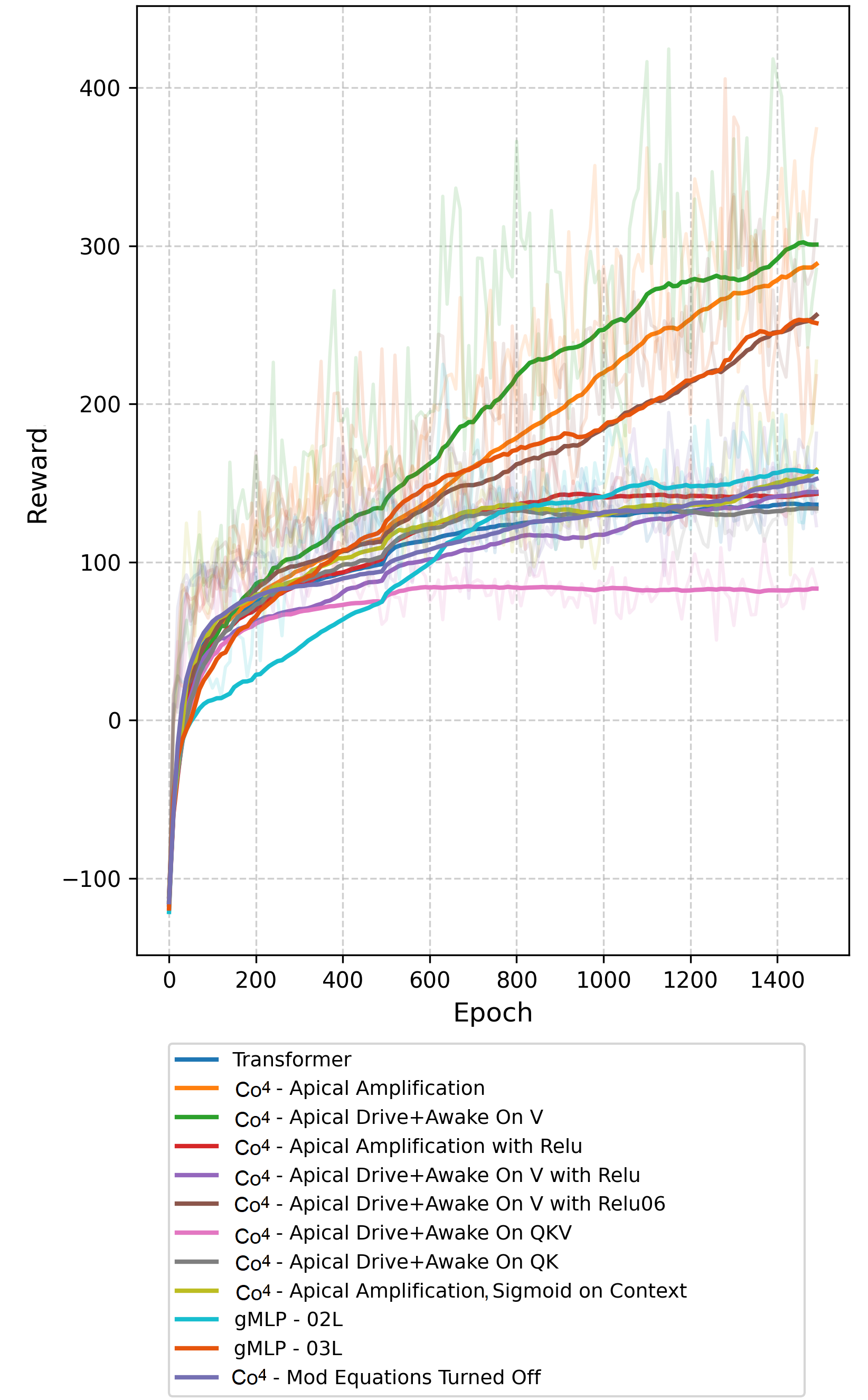}
\caption{\textbf{Training performance in the complex vision-guided quadrupedal locomotion environment.}
Running average reward during training for the Transformer and gMLP baselines and
Co$^4$ variants implementing different forms of apical
modulation \cite{adeel2026scalable,graham2025context}. Performance differs
substantially across modulation conditions. Disabling the Co$^4$ modulation
equations provides a control for assessing the functional contribution of
contextual modulation.
}
\label{fig:co4_ablation_2}
    \vspace{-0.2cm}

\end{figure}

\begin{figure}
\centering
\includegraphics[
    trim=0cm 0cm 0cm 0cm,
    clip=true,
    width=1\linewidth
]{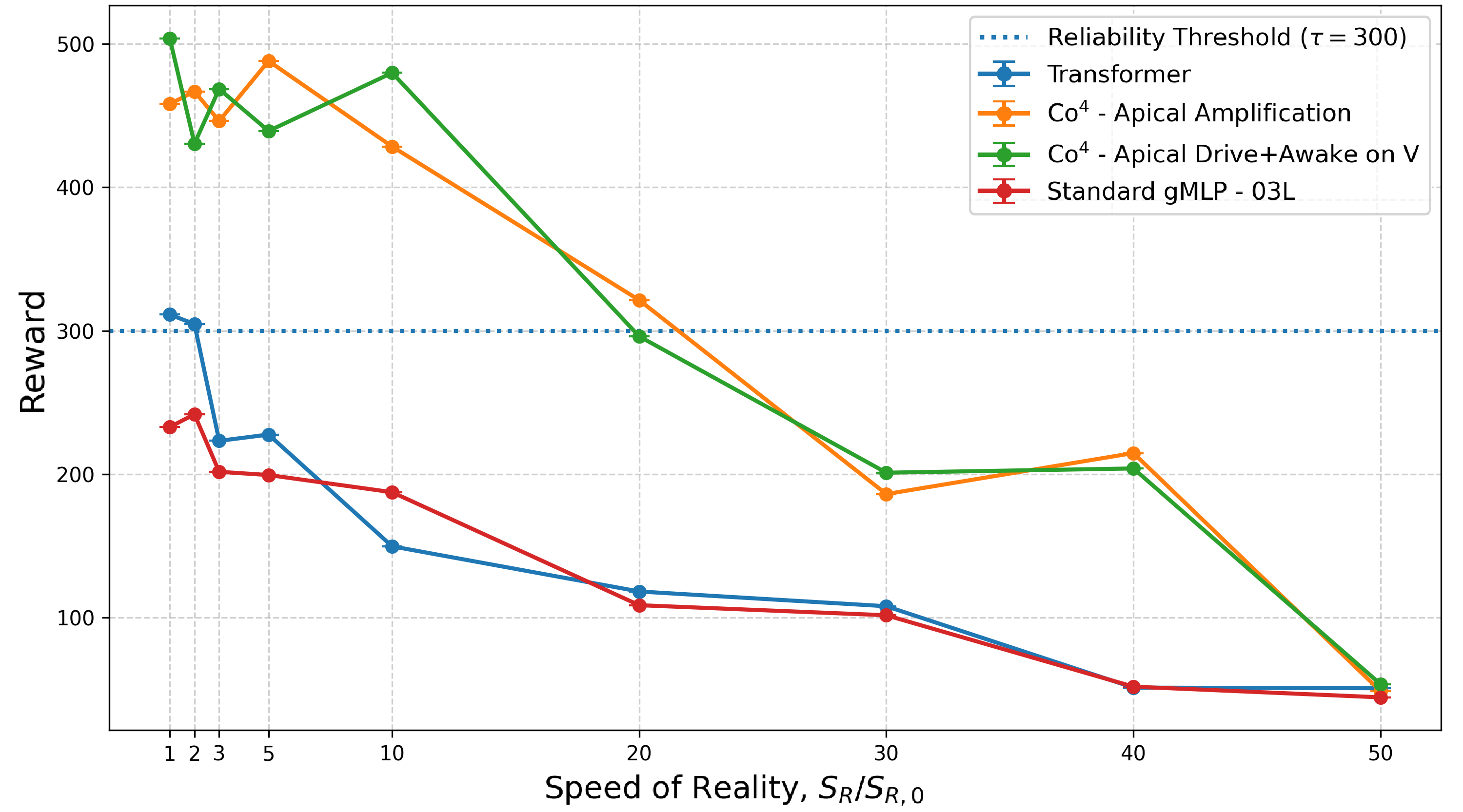}
\caption{
\textbf{{Scaling $S_C^{\mathrm{eff}}$ in the complex vision-guided quadrupedal locomotion task.}}
Reward is shown as $S_R$ increases from the training baseline
$S_{R,0}=1$ to $S_R=50$ without retraining. The dashed line denotes the
predefined reliability threshold $\tau=300$. Although the three-layer gMLP performed comparably to the two-layer \(Co^4\)
during training, it showed substantially weaker generalisation as \(S_R\) increased. Further architectural optimization, including increasing the number of layers, is beyond the scope of the present general challenge. }
\label{fig:quadruped_complex_sc}
\end{figure}
\subsubsection{Complex Locomotion Task}
Figure 6 shows training performance in the
complex vision-guided quadrupedal locomotion environment at $S_{R,0}=1$. The Transformer and $Co^4$ architectures each consist of two layers and four attention heads, whereas the gMLP architecture is evaluated with both two-layer and three-layer configurations. The Co$^4$-AA and Co$^4$-ADA achieve the highest running-average rewards over training,
whereas the Transformer, gated-MLP, and several $Co^4$ ablations and
alternative modulation schemes converge to lower reward levels. In particular, disabling the
Co$^4$ modulation equations \cite{adeel2026scalable, graham2025context} removes much of the performance advantage
observed for the better-performing Co$^4$ variants. These results
support the functional importance of contextual modulation for learned task
performance.

Importantly, performance at $S_R=1$ alone does not establish how well an
agent has captured the dynamics required to remain reliable when its access
to current reality is reduced. The subsequent $S_R$ test therefore evaluates
how far each trained system can be taken beyond its baseline observation
condition, without retraining, before its behavioural reliability criterion
is no longer satisfied.

\begin{table}[t]
\centering
\caption{Performance in the complex vision-guided quadrupedal locomotion environment as $S_R$ increases. Values are mean $\pm$ standard deviation.}
\label{tab:complex_locomotion}
\resizebox{\columnwidth}{!}{%
\begin{tabular}{ccccc}
\toprule
$\mathbf{S_R}$ 
& $\mathbf{Co^4}$-ADA 
& $\mathbf{Co^4}$-AA 
& \textbf{Transformer} 
& \textbf{gMLP} \\
\midrule
1  & 503.8 $\pm$ 28.6 & 458.0 $\pm$ 27.5 & 311.2 $\pm$ 18.7 & 232.8 $\pm$ 14.0 \\
2  & 430.4 $\pm$ 26.6 & 466.8 $\pm$ 28.0 & 304.7 $\pm$ 18.3 & 241.8 $\pm$ 13.8 \\
3  & 468.5 $\pm$ 28.6 & 446.2 $\pm$ 26.8 & 223.2 $\pm$ 13.4 & 201.7 $\pm$ 12.0 \\
5  & 439.2 $\pm$ 27.2 & 488.0 $\pm$ 29.3 & 227.6 $\pm$ 13.7 & 199.4 $\pm$ 11.1 \\
10 & 479.8 $\pm$ 27.2 & 428.3 $\pm$ 25.7 & 149.7 $\pm$ 9.0  & 187.4 $\pm$ 8.0  \\
20 & 302.0 $\pm$ 18.5 & 321.3 $\pm$ 19.3 & 118.2 $\pm$ 7.1  & 108.7 $\pm$ 6.8  \\
30 & 201.0 $\pm$ 11.7 & 186.1 $\pm$ 11.2 & 108.0 $\pm$ 6.5  & 101.7 $\pm$ 3.6  \\
40 & 204.0 $\pm$ 12.1 & 214.7 $\pm$ 12.9 & 51.2 $\pm$ 3.1   & 51.9 $\pm$ 4.6   \\
50 & 53.7 $\pm$ 3.2   & 48.8 $\pm$ 2.9   & 50.8 $\pm$ 3.0   & 44.4 $\pm$ 2.2   \\
\bottomrule
\end{tabular}%
}
\end{table}

\begin{figure}
\centering
\includegraphics[
    trim=0cm 0cm 0cm 0cm,
    clip=true,
    width=1\linewidth
]{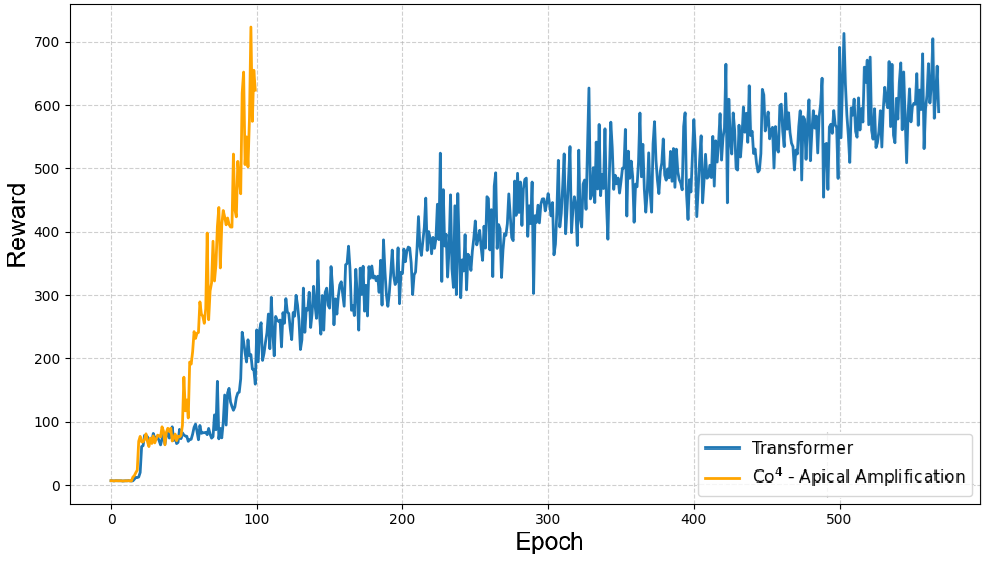}
\caption{
\textbf{Learning curves for the car-racing task.}}
\label{fig:car_racing_learning}
\end{figure}

\begin{figure}
\centering
\includegraphics[
    trim=0cm 0cm 0cm 0cm,
    clip=true,
    width=1\linewidth
]{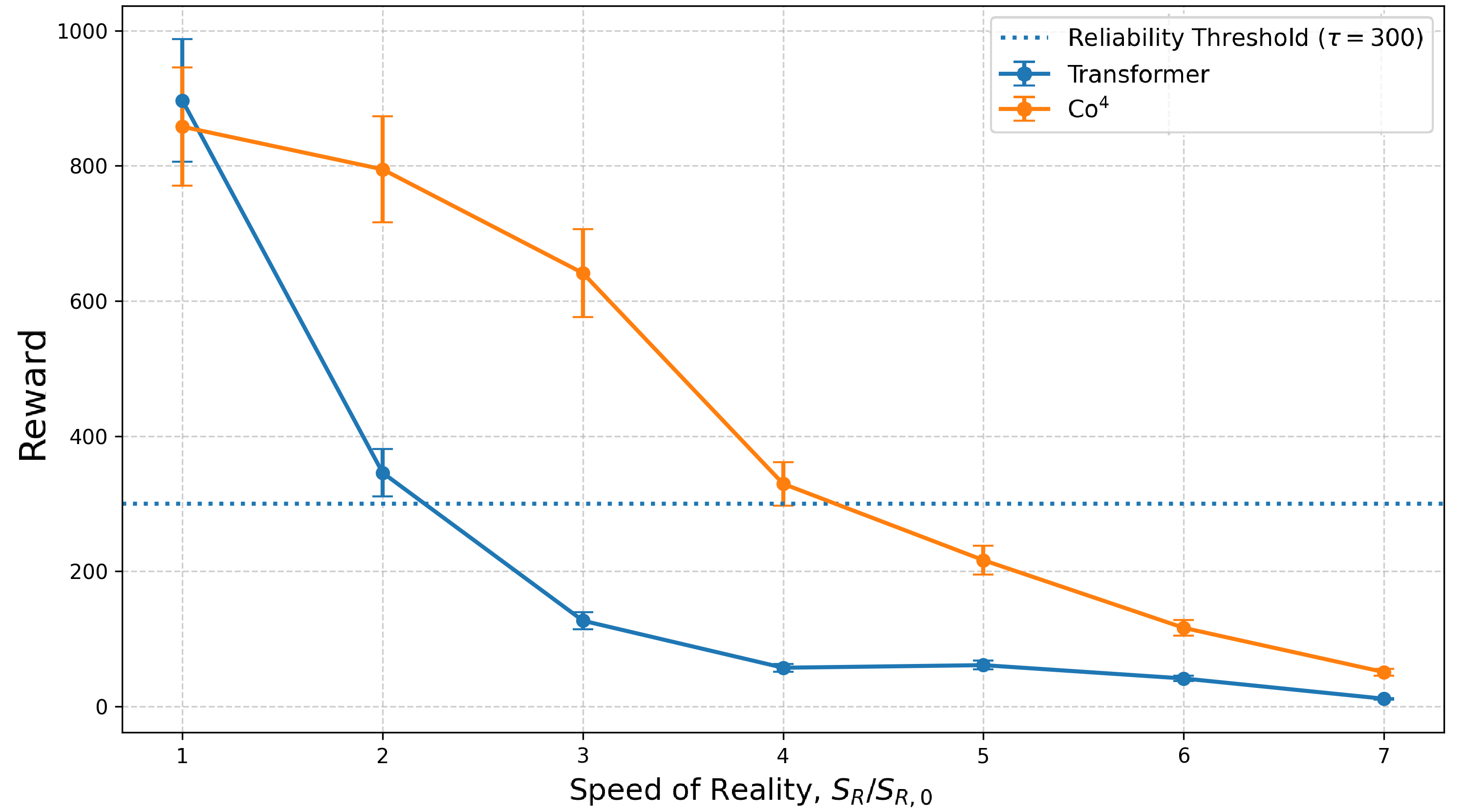}
\caption{
\textbf{Scaling the Speed of Coherent Cognition in the car-racing task.}
Mean reward for the Transformer baseline and Co$^4$ as $S_R$ increases
from $1$ to $7$, without retraining and with computational resources held
fixed. The dashed line indicates the predefined reliability threshold,
$\tau=300$. The Transformer remains reliable through $S_R=2$ and Co$^4$
through $S_R=4$, yielding $S_C^{\mathrm{eff}}=2$ and
$S_C^{\mathrm{eff}}=4$, respectively.
}
\label{fig:car_racing_sc}
\end{figure}
Figure 7 and Table 2 show the resulting reward as $S_R$
increases without retraining. The Transformer baseline maintains reliable behaviour through $S_R=2$ and
first falls below the reliability threshold at $S_R=3$, yielding
$S_{C,\mathrm{Transformer}}^{\mathrm{eff}}=2$. Co$^4$ remains reliable
through $S_R=20$ and first falls below threshold at the next tested operating
point, $S_R=30$, yielding $S_{C,\mathrm{Co}^4}^{\mathrm{eff}}=20$.
Under this predefined criterion and the tested operating points, Co$^4$
therefore extends the observed continuously reliable range of $S_R$ by a
factor of ten relative to the Transformer baseline under the tested operating points.

\subsection{Car Racing}
The car-racing models are also trained only at $S_{R,0}=1$. For this experiment, reliable behaviour is defined by the predefined reward
threshold $\tau=300$. The Transformer and $Co^4$ architectures each consist of two layers and four attention heads. Figure 8 depicts the learning curves of the two models. The $Co^4$–AA model reaches its peak reward after approximately 100 iterations, whereas the Transformer requires nearly 500 iterations to achieve a comparable reward level. This indicates substantially faster convergence for the $Co^4$–AA.

\begin{table}
\centering
\caption{Performance in the car racing task as $S_R$ increases. Values are mean $\pm$ standard deviation}
\label{tab:car_racing_speed}
\small
\begin{tabular}{ccc}
% \hline
\textbf{$S_R$} & \textbf{Co$^4$} & \textbf{Transformer} \\
% \hline
$1$ & $857.89 \pm 87.52$ & $896.71 \pm 91.02$ \\
$2$ & $794.74 \pm 78.28$ & $345.90 \pm 35.02$ \\
$3$ & $641.25 \pm 64.77$ & $126.97 \pm 12.42$ \\
$4$ & $329.49 \pm 32.16$ & $57.34 \pm 5.93$ \\
$5$ & $216.50 \pm 21.11$ & $61.18 \pm 6.48$ \\
$6$ & $116.39 \pm 11.46$ & $41.45 \pm 4.02$ \\
$7$ & $50.88 \pm 5.06$   & $11.48 \pm 1.16$ \\
% \hline
\end{tabular}
\end{table}

Figure 9 and Table 3 show the resulting reward as $S_R$
increases without retraining. The Transformer baseline
satisfies the reliability criterion through $S_R=2$ and first falls below
threshold at $S_R=3$, yielding
$S_{C,\mathrm{Transformer}}^{\mathrm{eff}}=2$. Co$^4$ remains reliable
through $S_R=4$ and first falls below threshold at $S_R=5$, yielding
$S_{C,\mathrm{Co}^4}^{\mathrm{eff}}=4$. Co$^4$ therefore extends the
observed continuously reliable range of $S_R$ by a factor of two relative
to the Transformer baseline under the tested operating points.

% \section{Effective mind–reality ratio}

% To relate the experimentally estimated $S_C^{\mathrm{eff}}$ to the conceptual mind–reality ratio, we define the effective ratio $\gamma^{\mathrm{eff}}=S_C^{\mathrm{eff}}/S_R$. Tables~\ref{tab:gamma_eff_locomotion} and~\ref{tab:gamma_eff_racing} report $\gamma^{\mathrm{eff}}$ across all tested operating points for the locomotion and car-racing experiments, respectively. By construction, $\gamma^{\mathrm{eff}}=1$ at the highest tested $S_R$ through which the predefined reliability criterion is continuously satisfied, while values above and below 1 indicate operating points below and beyond this empirical boundary. This effective ratio provides a normalized description of the observed reliability margin and is distinct from the theoretical ratio $\gamma=S_C/S_R$.

% Because $S_R$ is evaluated at discrete operating points, the precise
% continuous reliability boundary is not identified. The transition is
% localized between $S_R=2$ and $S_R=3$ for the Transformer and between
% $S_R=4$ and $S_R=5$ for Co$^4$. The full reward--$S_R$ relationship
% should therefore be considered alongside the scalar $S_C^{\mathrm{eff}}$,
% which depends on both the predefined reliability criterion and the tested
% operating points.

% These results establish a difference in the observed behavioural reliability
% boundary. We hypothesize that verification before propagation contributes
% to the higher $S_C^{\mathrm{eff}}$ observed for Co$^4$; however, this
% comparison does not isolate that mechanism from other architectural
% differences.

\section{Conclusion and Future Directions}
An important next step is to apply this challenge to
state-of-the-art AI and emerging world models
\cite{ha2018world, ding2025understanding, lecun2022path, maes2026leworldmodel},
as well as computational frameworks such as Active Inference
\cite{friston2010free}.
We hypothesise that the superior reliability demonstrated
by $Co^4$ may arise from an enhanced
capacity for grounded imagination
\cite{Phillips2024cellular, graham2026inhibitory, adeel2026scalable, adeel2025beyond}.
While the present proof of concept demonstrates behavioural
advantages consistent with this hypothesis, the underlying mechanisms of grounded imagination will be
investigated explicitly in future work involving scaled
world models.
Overall, for reliable AI, the scaling principle is simple:
\textit{scale $S_C$, not merely FLOPs, parameters, and data.}

\section{Impact}
Improving the epistemic reliability of increasingly capable AI systems is essential for their safe deployment \cite{dalrymple2024towards}. By challenging existing AI paradigms and introducing a new principle for scaling AI, this work offers a potential biologically inspired foundation for scaling justified trust in AI systems, with broader benefits for society \cite{adeel2026cellular}.

\section{Acknowledgments}
We acknowledge funding from the Advanced Research + Invention Agency (ARIA)
through the Nature Computes Better Opportunity Seeds programme. We thank
Professor Bill Phillips, Professor Leslie Smith, and Professor Bruce Graham
from the University of Stirling; Dr James Kay from the School of Mathematics,
University of Glasgow; Dr Lucy S. Petro from the School of Psychology and Neuroscience,
University of Glasgow; Professor Johan Frederik Storm, Professor of Neurophysiology at the University of Oslo, and Professor Hadi Larijani from Glasgow Caledonian University. We also acknowledge ChatGPT for
assistance with proofreading and coding support.

\appendix

\section{Related Work}
Prior work \cite{sharma2017learning, biedenkapp2021temporl, tallec2019making, ramstedt2019real, ramstedt2020reinforcement, anokhin2025handling, tessler2019action, zhang2026world, li2019robust, padakandla2021survey, kirk2023survey, xiao2020thinking, zhang2020robust, tobin2017domain} has studied several neighbouring aspects of temporal robustness in reinforcement learning, including observation and action delays, temporal abstraction and adaptive action duration, robustness to changes in control frequency and temporal discretization, and generalisation under environment or dynamics shifts. These directions address important forms of temporal and distributional variation, but differ from the evaluation considered here. In the proposed challenge, agents are trained exclusively at a baseline Speed of Reality, \(S_{R,0}=1\), and evaluated without retraining or adaptation as \(S_R\) is systematically increased while computational resources remain fixed. Thus, increasing \(S_R\) progressively enlarges the observation–reality gap beyond that experienced during training. The intervention is therefore not simply frame skipping, a change in sensory appearance or physical parameters, or training an agent to compensate for a known delay or altered control frequency. Instead, it tests whether task-relevant dynamics learned at the baseline support reliable action across increasingly distant temporal relationships that were not encountered during training. The resulting \(S_R\) sweep provides a model-agnostic stress test of zero-shot temporal generalisation under increasingly rapid task-relevant environmental change, complementing existing delay, control-frequency, robustness, and OOD evaluation paradigms.

\section{Experimental details} For quadrupedal locomotion, a single image is used to compute each output. Covering more than 50\% of the track is considered reliable behaviour. The robot is rewarded according to multiple objectives, ranging from keeping its body above the ground (i.e., remaining upright) to maintaining high velocities along the \(x\), \(y\), and \(z\) axes, and is penalised for contact with obstacles. The mean and standard deviation are computed over 10 runs. The simulation setup remains the same as that presented in \cite{yang2021learning}. 

For car racing, each training input consists of the current image frame together with the previous three frames. Each episode contains a maximum of 1000 frames while the car remains on the track. The car receives a reward for visiting each new track tile during training and is penalised for other behaviours. Reliable behaviour is defined as achieving at least 50\% of the task, with the corresponding reward score used to define the reliability threshold. The mean and standard deviation are computed over 10 runs. The simulation setup remains the same as that presented in \cite{tang2021sensory, ha2018world}.

\subsection{Architectural and Computational Comparison}
$Co^4$ and the Transformer baseline are matched in parameter
count. Unlike conventional full attention, $Co^4$ employs
contextual modulation to distinguish valid and relevant
from invalid or irrelevant $Q$-$K$-$V$ token representations
before downstream processing. Once contextual verification
has been performed, the downstream readout can be selected
according to architectural and computational requirements,
including attention, pruning, gating, or MLP-based processing.

In the present proof-of-concept implementation, attention
is restricted to the top-$k$ contextually relevant tokens,
yielding an attention-related computational complexity of
$O(N+k^2)$, where $N$ is the number of input tokens,
$k\ll N$, and $k\leq\sqrt{N}$. Thus, the implementation combines contextual verification with computationally selective attention, achieving substantially improved behavioural reliability with matched parameter counts and reduced computational FLOPs relative to the Transformer baseline.  Further details are provided in
\cite{adeel2026scalable, adeel2025beyond}.

Code and project resources are available through the TREND Project:
\url{https://github.com/ARIA-Funded-TREND/IHMS}.

\bibliographystyle{IEEEtran}
\bibliography{NATURE.bib}

\end{document}